\documentclass{article}

\usepackage{PRIMEarxiv}

\usepackage[utf8]{inputenc} 
\usepackage[T1]{fontenc}    
\usepackage{hyperref}       
\usepackage{url}            
\usepackage{booktabs}       
\usepackage{amsfonts}       
\usepackage{nicefrac}       
\usepackage{microtype}      
\usepackage{lipsum}
\usepackage{fancyhdr}       
\usepackage{graphicx}       
\graphicspath{{media/}}     
\usepackage{amsmath}
\usepackage{mathrsfs}
\usepackage{tabularx}

\title{Carbon Market Simulation with Adaptive Mechanism Design}

\author{
Han Wang$^1$,
Wenhao Li$^1$,
Hongyuan Zha$^1$,
Baoxiang Wang$^{1,2}$\\
$^1$The Chinese University of Hong Kong, Shenzhen\\
$^2$Vector Institute\\
xwanghan@gmail.com,
\{liwenhao,zhahy,bxiangwang\}@cuhk.edu.cn
}

\begin{document}
\maketitle

\begin{abstract}
    A carbon market is a market-based tool that incentivizes economic agents to align individual profits with the global utility, i.e., reducing carbon emissions to tackle climate change. 
    \textit{Cap and trade} stands as a critical principle based on allocating and trading carbon allowances (carbon emission credit), enabling economic agents to follow planned emissions and penalizing excess emissions. 
    A central authority is responsible for introducing and allocating those allowances in cap and trade. 
    However, the complexity of carbon market dynamics makes accurate simulation intractable, which in turn hinders the design of effective allocation strategies.
    To address this, we propose an adaptive mechanism design framework, simulating the market using hierarchical, model-free multi-agent reinforcement learning (MARL). 
    Government agents allocate carbon credits, while enterprises engage in economic activities and carbon trading. 
    This framework illustrates agents' behavior comprehensively. 
    Numerical results show MARL enables government agents to balance productivity, equality, and carbon emissions. 
    Our project is available at \url{https://github.com/xwanghan/Carbon-Simulator}.
\end{abstract}


\section{Introduction}
    
    Climate change has emerged as a pressing worldwide concern ~\cite{45b2d85e-24d3-334f-b350-d7218b81a7bb}, significantly imperiling global ecosystems, economic systems, and sociopolitical stability. 
    The United Nations reports that in developing regions, one in ten individuals subsists on less than US\$ $1.90$ daily ~\cite{unpoverty}, with $2.2$ billion people deprived of access to safely managed potable water resources ~\cite{unwater}. 
    The burgeoning climate crisis amplifies these challenges, as worldwide temperature escalations provoke droughts and rising sea levels, exacerbating famines and enhanced forced displacements ~\cite{unclimate}.
    
    In $2016$, $196$ nations endorsed the Paris Agreement to mitigate climate change collaboratively. 
    However, pursuing transnational environmental targets often conflicts with short-term interests, requiring mechanisms to reconcile local and international objectives ~\cite{paris}. 
    Carbon markets exemplify such mechanisms ~\cite{protocol1997kyoto,wara2007global,zhou2019carbon}, incentivizing economic agents to curb emissions. 
    The \textit{cap and trade} format, predominant in carbon markets, involves \textit{allocating} and \textit{trading} allowances ~\cite{goulder2013carbon,schmalensee2017design,ZHOU201647,Carbon_trade}.
    Economic agents must possess sufficient allowances to offset emissions or face penalties for surplus emissions.
    The \textit{cap and trade} system sets a predetermined limit on allowances within an economy, with a central authority introducing and allocating allowances based on specified objectives. 
    While this policy helps balance efficiency and fairness, determining the optimal allocation remains challenging in general economic contexts. 
    The high-dimensional dynamics of the carbon market, influenced by rational, self-interested, and far-sighted economic agents, lead to market simulation reliance on models like CGE (computable general equilibrium) ~\cite{hubler2014designing,tang2016designing,bi2019impact} or ABM (agent-based modeling) ~\cite{tang2017carbon,zhou2016multi,de2021carbon} frameworks, 
    employing simplifying assumptions that are arduous to validate, such as production and trading behaviors.
    
    Given the unique nature of the carbon market, we integrate the \textit{AI Economist} ~\cite{zheng2022ai,trott2021building} to simulate market dynamics. 
    Our adaptive mechanism design framework, employing hierarchical, model-free MARL, mimics the carbon market. 
    Lower-level enterprise agents engage in realistic economic activities, such as emitting carbon dioxide, trading emission credits, and investing in emission reduction projects. 
    Higher-level government agents analyze diverse allocation strategies to achieve balanced efficiency and fairness, leading to significant carbon emission reductions. 
    The framework demonstrates the conduct of rational, self-interested, and far-sighted agents within the carbon market. 
    We emphasize that our approach is not a simple transfer of the AI-Economist from taxation to carbon credit allocation. 
    Simulating the carbon market is challenging due to limited data, fluctuating regulations, and non-market factors.
    
    To validate our simulator, we conducted comparisons with several widely adopted indicator allocation approaches at the firm level~\cite{ZHOU201647}.
    The simulation results indicate reasonable action responses by enterprise agents to these allocation policies. 
    Additionally, numerical findings demonstrate that government agents, through MARL, effectively discover allocation policies capable of balancing productivity, equity, and carbon emissions. 
    Our primary contributions encompass: 
    \textbf{1)} We propose a systematic carbon market simulator featuring carbon credits allocation and trading, and achieve realistic carbon economy simulation based on hierarchical, model-free MARL.
    \textbf{2)} We implement several widely adopted indicator allocation approaches at the firm level as baselines.
    \textbf{3)} We observe that learning-based allocation policies possess the potential to effectively balance productivity, equity, and carbon emissions.

\begin{figure*}[htb!]
    \centering
    \includegraphics[width=0.93\linewidth]{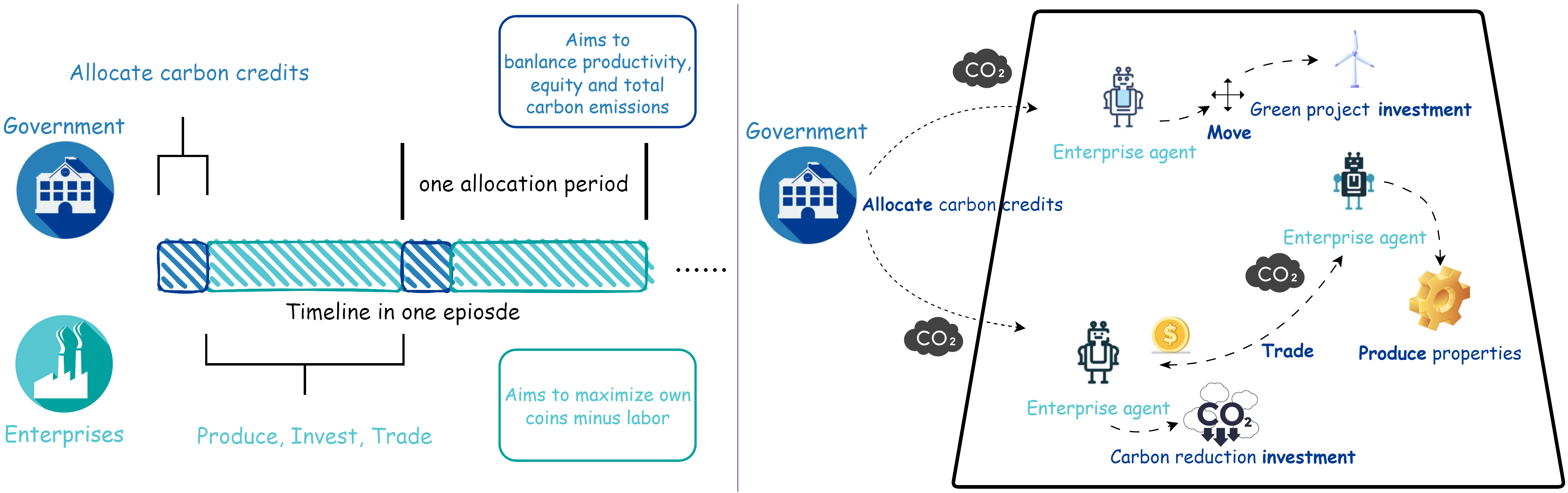}
    \caption{Simulator structure. \textbf{Left:} One episode is divided into several periods, and in each periods, the government firstly acts to allocate carbon credits; the remaining of time, enterprises do their economics activities. \textbf{Right:} Enterprises' economics activities are modeled in a Gather-Trade-Build game; in this grid map, they can produce properties (build) to get coins, can move which can gather carbon credits and increase community's total power (green project investment), do carbon reduction invest action to reduce the carbon emission level (carbon reduction investment), also trade carbon credits and coins with each other.}
    \label{fig:Structure}      
\end{figure*}

\section{Preliminaries and Related Works}\label{sec:Background}

\subsection{Carbon Market and Cap and Trade}

A carbon market can manifest at various scales, from local to global. 
This paper primarily focuses on regional carbon markets, which involve the participation of the government and many enterprises. 
Carbon market allowances can be classified into \textit{mandatory} and \textit{voluntary} types, with the former being strictly regulated by the government. 
At the same time, the latter encourages enterprises to decrease their emissions through investment in carbon credits generated by government-certified projects, facilitating emissions avoidance, reduction, or removal from the atmosphere. 
The most prevalent form of carbon markets is the cap and trade system, encompassing a predefined cap on total allowances within the economy at a specific time. 
A central authority or government manages the introduction of these allowances, with the cap value being established to align with their objectives. 
This mechanism ensures an accurately targeted level of emissions.

Carbon credits can be introduced into the market through $3$ primary means. 
The initial method, termed \textit{free allocation}, entails the government determining the number of free credits to allocate to enterprises based on factors such as the enterprise's size and historical emissions. 
Another approach involves the government selling allowances in an \textit{auction market}. 
The final method incorporates \textit{certified projects}, enabling enterprises to garner additional allowances by undertaking and completing energy conservation and emissions reduction projects, typically facilitated by technological investment and advancement. 
This paper exclusively focuses on free allocation and government-certified projects.

In the context of free allocation, a pivotal decision confronting the government is the establishment of both the overall volume of allowances for a given period and the distribution of these allowances among enterprises within distinct regions ~\cite{baer2013greenhouse,bohringer2005design,yi2011can}. 
Allocation policies can be primarily divided into four approaches: indicator, optimization, game theoretic, and hybrid approaches. 
Each approach possesses strengths and weaknesses in terms of productivity and equity. 
The indicator approach is the most frequently employed in practice; 
however, consensus on indicator selection remains elusive, as differing indicator methodologies can yield substantially varied allocation outcomes and are challenging to validate ~\cite{ZHOU201647}. 
In our simulations, we utilized the indicator policy as a baseline to compare against the MARL-based allocation policy.

The paramount component within the cap and trade is carbon trading ~\cite{protocol1997kyoto,gonzalez2011importance,Carbon_trade}. 
Enterprises that exceed allotted emissions limits can elude penalties by purchasing allowances from entities possessing surplus allowances. 
This paper simulates carbon trading using a \textit{bid-and-ask} auction mechanism, facilitating allowance-based transactions between enterprises, analogous to the AI Economist~\cite{zheng2022ai}.

\subsection{Carbon Market Simulated Models}

The prevailing quantitative simulator encompass the computable general equilibrium model(CGE) ~\cite{hubler2014designing,tang2016designing,bi2019impact} and the agent-based model(ABM) ~\cite{tang2017carbon,zhou2016multi,de2021carbon,yu2020modeling}.
CGE models are economic frameworks that obtain empirical data as input to emulate economic structures and the behavioral response of economic entities as accurately as feasible to examine potential influences of varied policies and other disturbances. 
CGE and dynamic CGE models have been employed extensively to broadly explore climate policies, specifically, the carbon market ~\cite{edwards2001allocation,tang2016designing}.
Moreover, multi-agent-based models have been utilized to replicate distinct industries' behavior at a national scale ~\cite{tang2017carbon,yu2020modeling,de2021carbon}. 
~\cite{zhou2016multi} propose a multi-agent simulation framework to emulate regional emissions trading systems and assess numerous regulatory policies and carbon auction regulations.
Additionally, other multi-agent-oriented models have been applied for exploring specific aspects of emissions trading schemes, such as prospective carbon auction rules ~\cite{esmaeili2022auction} or international transportation patterns ~\cite{mizuta2008agent}. 
Rafieisakhaei et al. ~\cite{rafieisakhaei2016modeling} introduces models targeting the EU ETS and the global oil market, subsequently examining the interrelation between these components.

\subsection{Machine Learning for Economic Design}

In the domain of economic design, contemporary investigations have chiefly centered on automated mechanism design ~\cite{conitzer2002complexity} and auction design ~\cite{dutting2015payment}, with the primary aim being the augmentation of market efficacy and resource allocation via the employment of machine learning methodologies. 
Historically, agents within such research contexts have functioned as static entities devoid of learning capabilities.
However, recent scholarly contributions have begun to incorporate MARL for the examination of security games ~\cite{wang2019deep} on Stackelberg equilibrium, as well as the investigation of resource allocation games ~\cite{balduzzi2019open}. 
Furthermore, certain studies have extended the application of RL by enabling agents to modify their conduct in adherence to novel regulatory frameworks, thereby intensifying the refinement of market design ~\cite{tang2017reinforcement, shen2020reinforcement} and aligning with the "Positronic Economist" paradigm ~\cite{thompson2017positronic}.
Moreover, AI economists ~\cite{zheng2020ai} have successfully optimized taxation policies within a two-level simulation environment, surpassing the effectiveness of extant real-world tax frameworks.

\subsection{Multi-Agent Reinforcement Learning}

In the simulated environment under investigation, numerous enterprise agents adapt in response to distinct government policy interventions. 
These agents' training process is grounded in the principles of MARL. 
Owing to the complexity of environment state transitions, which encompass not only stochastic environmental factors but also the actions of all agents involved, a single agent's perspective may inadvertently conflate environmental randomness with the behavior of other agents, ultimately leading to improper agent updates ~\cite{claus1998dynamics}.
A growing body of recent research in the domain of MARL has been devoted to improving multi-agent learning in non-stationary environments ~\cite{laurent2011world}, employing strategies such as reward sharing ~\cite{rashid2020monotonic, sunehag2017value} or state information sharing ~\cite{lowe2017multi, yu2022surprising} amidst agents. 
This paper adopts an alternative approach to bolster the learning process: sharing agent parameters ~\cite{rashid2020monotonic} while precluding the exchange of individual agent information.


\section{Carbon Market Modeling}\label{sec:environment}

In this section, we will present an approach to conduct simulations of carbon markets within the context of model-free MARL.
The MARL framework exhibits a hierarchical structure, consisting of the higher-level government RL agent and lower-level enterprise RL agents. 
Consequently, it is referred to as a hierarchical model-free MARL framework, which has also been known as a manager-worker architecture ~\cite{shu2018mrl,ma2020feudal} in existing literature. 
Section~\ref{sec:hmarl} will provide an in-depth mathematical representation of this framework, while Sections~\ref{sec:higher-government} and \ref{sec:lower-enterprise} will elucidate the instantiated models for government and enterprise RL agents, respectively. We use notations reference from RL and economics, see Table \ref{tab:Notations}.

\subsection{Hierarchical Multi-Agent RL Framework}\label{sec:hmarl}

\begin{table}[htb!]
\centering
\label{tab:Notations}
\resizebox{0.7\columnwidth}{!}{%
\begin{tabular}{lllll}
\cline{1-4}
$t$              & \multicolumn{1}{l|}{timestep}                     & $\pi_{g}$ & government allocation policy &  \\
$i, j, k$        & \multicolumn{1}{l|}{enterprise agent indices} & $p$       & punishment                   &  \\ \cline{3-4}
$g$              & \multicolumn{1}{l|}{government agent index}   & $z$       & income                       &  \\ \cline{1-2}
$\theta$, $\phi$ & \multicolumn{1}{l|}{model weights}            & $l$       & labor                        &  \\
$h$              & \multicolumn{1}{l|}{hidden state}             & $c^l$     & labor-income coefficient     &  \\ \cline{1-4}
$s$              & \multicolumn{1}{l|}{state}                    & $u$       & utility                      &  \\
$o$              & \multicolumn{1}{l|}{observation}              & $swf$     & social welfare               &  \\
$a$              & \multicolumn{1}{l|}{action}                   & $prod$    & productivity                 &  \\
$r$              & \multicolumn{1}{l|}{reward}                   & $eq$      & equality                     &  \\
$\pi$            & \multicolumn{1}{l|}{policy}                   & $ee$      & excess carbon emissions      &  \\
$\gamma$         & \multicolumn{1}{l|}{discount factor}          & $c^e$     & economy-climate coefficient  &  \\
$\mathscr{T}$    & \multicolumn{1}{l|}{state transition}         &           &                              &  \\ \cline{1-4}
x                & enterprise attributes                         & $x^q$     & carbon emission credit       &  \\
$x^c$            & coin                                          & $x^l$     & carbon emission level        &  \\ \cline{1-4}
                 &                                               &           &                              &  \\
                 &                                               &           &                              & 
\end{tabular}%
}
\caption{Notations for our Hierarchical Multi-Agent RL Framework.}
\end{table}

On the one hand, the issues encountered by higher-level government agents in the carbon market (primarily referring to carbon allowances allocation within this paper) can be modeled as a standard Markov decision process(MDP) ~\cite{bellman1957markovian} $\mathcal{M}_{h} = (\mathcal{S}, \mathcal{A}, \mathcal{P}, \mathcal{R}, \gamma)$, given the fixed lower-level enterprise agents.
Given states $s, s^{\prime} \in \mathcal{S}$ and an action $a \in \mathcal{A}$, the transition probability function is expressed as $\mathcal{P}\left(s^{\prime} \mid s, a\right): \mathcal{S} \times \mathcal{A} \times \mathcal{S} \rightarrow[0, 1]$ and the reward function is defined by $\mathcal{R}(s, a, s^{\prime}): \mathcal{S} \times \mathcal{A} \times \mathcal{S} \rightarrow \mathbb{R}$. 
The discount factor is denoted as $\gamma \in (0, 1]$ and the policy is represented as $\pi: \mathcal{S} \times \mathcal{A} \rightarrow[0, 1]$.
For the timestep $t \in [1, T]$, the discounted return is identified by $G_t = \sum_{t^{\prime} = t}^T \gamma^{t^{\prime} - t} r_{t^{\prime}}$. 
The objective of the government agent is to determine a policy $\pi$ that maximizes $J = \mathbb{E}_{a_t \sim \pi(\cdot \mid s_t), s_{t + 1} \sim \mathcal{P}(\cdot \mid s_t, a_t)}[\sum_{t = 1}^T \gamma^{t - 1} r_t(s_t, a_t, s_{t + 1})]$.

On the other hand, the problems (production, trade, etc.) encountered by the lower-level enterprise agents in carbon markets can formally defined by a partially observable stochastic game(POSG) ~\cite{hansen2004dynamic} $\mathcal{G}_{l}:(\mathcal{S},\mathcal{A}_{i=1}^{|\mathcal{I}|},\mathcal{R}_{i=1}^{|\mathcal{I}|},\mathcal{P},\mathcal{O}_{i=1}^{|\mathcal{I}|},\mathcal{I})$, given a fixed higher-level government agent.
At each timestep $t\in T$, lower-level enterprise agent $i\in \mathcal{I}$ obtains an observation, $o_t^i = \mathcal{O}_i(s_t)$, where $\mathcal{O}_i(s_t)$ is emission or observation function. 
Then each agent $i$ execute an action $a_t^i \in \mathcal{A}_i$ simultaneously according to their historical-dependent policy $\pi_{i}:\mathcal{O}^1_i \times \mathcal{A}^1_i \times \cdots \times \mathcal{O}^t_i \times \mathcal{A}^t_i \rightarrow [0, 1]$.
After that, environment will return rewards $r_t^i \in \mathcal{R}_i(s_t, \boldsymbol{a}_t)$ to agent $i$, where $\boldsymbol{a}_t$ denotes the joint action of all agents.
Then state $s_t$ will transitions to the next state $s_{t+1}$ according to transition function $\mathcal{P}(s_{t+1}\vert s_t,\boldsymbol{a}_t)$. 
Each agent's objective is find an optimal policy $\pi$ to maximize the $\gamma$-discounted expected return $
\max_{\pi_i} \mathbb{E}_{a^i \sim \pi_i, \textbf{a}^{-i} \sim \pi_{-i}, s' \sim \mathcal{P}} \left[\sum_{t} \gamma^t r_t^i\right]$, where $\pi_{-i}$ means the joint policy of other agents.
Note that government and enterprises are make decisions in different timescales.

Subsequently, we instantiate $\mathcal{M}_h$ and $\mathcal{G}_l$, representing government and enterprises within the carbon market context, and provide definitions of state spaces, action spaces, and reward functions.
    
\subsection{Lower-Level: Enterprise Modeling}\label{sec:lower-enterprise}

The Gather-Trade-Build game, as proposed in AI-Economist ~\cite{trott2021building}, is employed to emulate enterprise economic behavior within the context of a carbon market. 
This two-dimensional grid world enables agents to traverse the grid, acquire resources, amass coins (representing profit) through resource usage in house construction endeavors, and partake in trading activities with other agents by exchanging resources for coins.
More specifically, enterprises are stochastically initialized on grid cells at the onset of each episode. 
Subsequently, these entities initiate actions at every timestep throughout an episode, excluding when the government intervenes (to be detailed later on). 
Upon the completion of each episode, a penalty equal to $ee*p$ is imposed on enterprises' coin holdings in response to surpassing their allotted carbon emission credits.

\paragraph{Reward Function.}
The objective of each enterprise is to maximize its utility which is defined as:
\begin{eqnarray}\label{Enterprise reward}
u(z,l) = \frac{z^{1-\eta}-1}{1-\eta} - c^l*l,   \quad\quad \eta = 0.23,
\end{eqnarray} where $z$ is income that the sum of enterprise‘s coins, and $l$ is also the cumulative labor during all previous time steps. 
And for $\eta$ is isoelastic coefficient, we use isoelastic utility function ~\cite{debreu1954representation} to model the income component of reward, it is a nonlinear curve with higher $\eta$ lower sensitivity to the income change.
During an episode, labor-income coefficient $c^l$ is changing, it follows equation 
\begin{eqnarray}\label{labor}
\alpha*(1-e^{t/\beta}),
\end{eqnarray} where $\alpha,\beta$ is constant and $t$ refers to time.

\paragraph{Skill.}
Enterprise agents' skills include $S$ and $Rc$, representing enterprise size and research and development (R\&D) capabilities.

\paragraph{Action Space.}
An episode is divided into several periods.
At the start of each period, each enterprise is allocated a certain number of carbon emission credits by the government agent. 
At each timestep within a period, enterprises must choose from a set of actions, including \texttt{Move}, \texttt{Produce}, \texttt{Invest}, \texttt{Trade} or \texttt{NO-OP} (no operation).

\texttt{Trade} action consumes one unit of labor. 
Enterprise agents can move to any of the four neighboring grids. 
Still, this action will be blocked by the edge of the grid world or other enterprises' properties (or houses). 
Suppose an enterprise agent moves to a grid already allocated a government-certified project. 
The enterprise will consume coins and labor and receive carbon emission credits. 
Once completed, the project will remain on the grid and influence the carbon emission process of all agents.

\texttt{Produce} action consumes one unit of labor, $x^l$ carbon emission credits, and provides the agent with $10S$ coins. This action is available if the current position in the grid environment is not occupied by the enterprise's property or a government-certified project. 
Additionally, when a large amount of carbon emission is created, it will randomly \textit{pollute} the value of nearby blank grids. 
This will introduce a discount on the coins received when the enterprise produces in such grids.

\texttt{Invest} action can decrease an enterprise's carbon emission level. 
This action is available when an enterprise has positive coins.
It consumes one unit of labor and $5/Rc$ coins.
The carbon emission level, denoted as $x^l$, is a continuously changing enterprise attribute, with a floating range between $0$ and $1$. 
At the beginning of the episode, all enterprises have their $x^l$ set to $1$. 
Through investments, an enterprise can change its $x^l$. 
The carbon emission level is defined as follows:
\begin{eqnarray}\label{invest}
x^l_i = Pc_i * (1-Gr),
\end{eqnarray} where $Pc_i$ is defined as $Pc_i = exp(-\delta * Rc_i * n_i^r)$ and $Gr$ is:
\begin{eqnarray}
Gr = \frac{n^p}{\sum_{j}^{I}(Pc_j*S_j)+n^p}.
\end{eqnarray} 
We use power consumption $Pc$ to represent the power consumed by an enterprise when producing $10$ coins. 
This value depends on the number of investments $n_i^r$, the enterprise's R \& D capabilities, and constant $\delta$.
The green rate $Gr$ is used to represent the proportion of green energy generated from government-certified projects in total energy. 
It decreases with a higher number of completed government-certified projects located in the grid world, denoted as $n^p$, or a lower total power consumption $\sum_{j}^{I}(Pc_j*S)$.
Additionally, we have set lower bounds to ensure that $x^l_i$ remains within a reasonable range and $Gr$ is non-negative. 
We have introduced delay, forgetting, and random failures in the count of investment actions $n_i^r$ to model actual carbon emission reduction investments.

\texttt{Trade} encompasses bidding and asking in carbon trading, similar to AI Economist.
The action space for \texttt{Trade} is twice the number of specified price levels. 
When an enterprise publishes a bid or ask request, that request remains active for a predetermined number of timesteps until it is matched with a higher-priced ask or a lower-priced bid. 
The range for the lifetime of a request should be smaller than the length of an episode because, at the start of each new episode, market requests are cleared.
Publishing a request in the market incurs a lower labor cost than other actions, encouraging trade activity. 
However, this action is subject to limitations. 
Enterprises can only publish a maximum allowed number of requests, and they must have a sufficient amount of coins and carbon emission credit to participate in trading.

\paragraph{Observation Space.}
The range of observation for enterprises is significantly smaller than that of the government.
Regarding position information, enterprises can gather information about their properties' positions and a limited square range of the grid world. 
Within this limited square range, enterprises can observe which grid areas are polluted by carbon emissions resulting from their production actions or purified by completing government-certified projects.
Enterprises will only know their skills and attributes but can access information about the average carbon emission level.
In carbon trading, enterprises can observe the embedded price history and all published requests, similar to what the government can monitor. 
Additionally, enterprises' observations will include self-published requests.


\subsection{Higher-Level: Government Modeling}\label{sec:higher-government}

\noindent\textbf{Reward Function.}
Government's objective is to maximize social welfare, which is defines as:
\begin{eqnarray}\label{government reward}
swf = prod * eq*exp(-c^e * ee),
\end{eqnarray} where $prod$ is defined as:
\begin{eqnarray}
prod = \sum_{i}(x_i^c),
\end{eqnarray}
and $eq$ defined from concept of $gini$ index:
\begin{eqnarray}
eq = 1- \frac{\sum_{i=1}^{N} \sum_{j=1}^{N} |x_{ci} - x_{cj}|}{2(N-1)\sum_{i=1}^{N}x_{ci}}, \quad\quad 0<eq<1.
\end{eqnarray}

\paragraph{Action Space.}
The government agent makes decisions at the first timestep of each period. Its action is represented by a multi-dimensional discrete vector. The first dimension specifies the proportion of the total credit for this period that will be allocated to the remaining unallocated carbon emission credits for all periods. The next $|\mathcal{I}|$ dimensions, where $|\mathcal{I}|$ refers to the number of enterprise agents, determine the allocation weight of carbon emission credits for each enterprise agent in each period.

The total credit available for the period is calculated based on the specified proportion of the period's full credit. The government then allocates 10\% of this credit to a government-certified project randomly placed on an empty grid. The remaining 90\% of the total credit is distributed to enterprises according to their respective allocation weights.

Each dimension of the action space has a size of 101, with the remaining action 0 used for masking at the rest of the timesteps in each period.

To ensure the proper functioning of a cap and trade, it is essential for the government agent not only to allocate carbon credits to each period and enterprise agent but also to establish the severity of penalties for exceeding emissions.
Thus, the $|\mathcal{I}|+2$-th dimension determines the severity of punishments.

Regarding different facets of government policy, the parameter that exerts the most significant influence on enterprise behavior is the punishment setting, as it directly impacts the utility of enterprises.
When the punishment is excessively high, enterprises will avoid exceeding the carbon emission credit at all costs, negatively affecting productivity and trade. 
Conversely, suppose the punishment is set too low. 
In that case, production actions will go unchecked, resulting in low equality and exceedingly high carbon emissions. 
Therefore, we have maintained a constant punishment for most of our experiments.
After each period, the carbon emission credits allocated to enterprises are reset while the government project remains intact.

\paragraph{Observation Space.}
The government possesses access to comprehensive information. 
Firstly, it can observe the entire grid world, including the positions of all enterprise agents, their properties' locations, and the government-certified project's location.
Next, concerning the private information of enterprises, the government can collect data on their skills and attributes. Additionally, real-time market information, including embedded price history and all published requests, is readily available. Furthermore, the government has access to information related to carbon emission credits, such as excess records and the total remaining credit.

\paragraph{Remark:}
Notably, in developing a carbon market model similar to AI-Economist, we opt for a selection of hyperparameters without referencing empirical data, relying solely on fundamental economic principles. 
This endows the simulation model with several advantages. 
Firstly, government and enterprise agents possess no prior knowledge of external markets or economic theory and lack any understanding of each other's behavioral patterns. 
Consequently, the simulator can optimize arbitrary social outcomes through hyperparameter configuration. 
Secondly, carbon market data is scarce and typically encompasses relatively small market scales. 
Our proposed simulator, by using appropriate hyperparameter settings, transcends the limitations of real-world data, enabling the examination of allocation strategies and the impact of economic activities under varying carbon market scales. 
Lastly, the simulator can validate the rationality of different economic theories by adjusting hyperparameters following various theoretical perspectives.


\begin{figure}[htb!]
    \centering
    \includegraphics[width=0.6\linewidth]{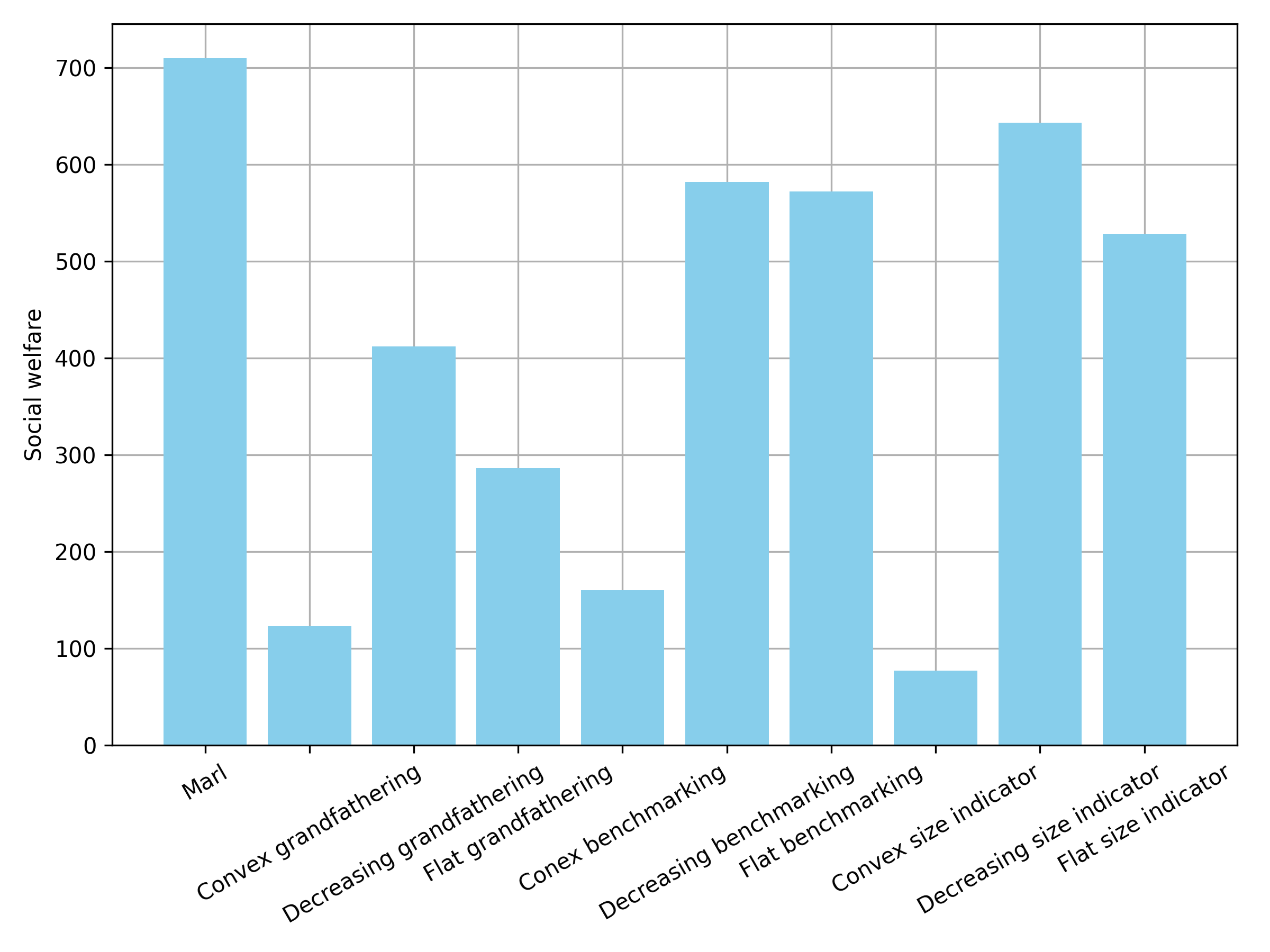}
    \caption{Quantitative results of different allocation policies.}
    \label{fig:bar}
\end{figure}   

\section{Enterprise Behavior Simulation}\label{sec:AP}

Upon the completion of carbon market modeling, we can employ (MA)RL to train the government (enterprise) agent(s), enabling them to exhibit behavior that closely resembles real-world scenarios, thereby achieving a realistic simulation of the carbon market. 
This section primarily focuses on the simulation of the critical entities in the carbon market, namely the behavior of enterprise agents.
Based on reward function we defined in Equation~(\ref{Enterprise reward}), we can obtain desired policy $\pi_i$ by maximizing total discounted reward:
\begin{eqnarray}
\max_{\pi_i}\mathbb{E}_{a^i \sim \pi_i, \textbf{a}^{-i} \sim \pi_{-i}, s' \sim \mathscr{T}}
\left[
\sum_t^T \gamma^t (u_{i,t} - u_{i,t-1})+u_{i,0}
\right].
\end{eqnarray} 
Like AI Economist, we use a strong baseline in MARL, the independent PPO~\cite{de2020independent}, to train adaptive response enterprise agents for government policy.
We incorporate $5$ enterprise agents into the carbon market in the simulation.

\subsection{Training Details}

\begin{figure}[htb!]
  \centering
  \includegraphics[width=0.6\linewidth]{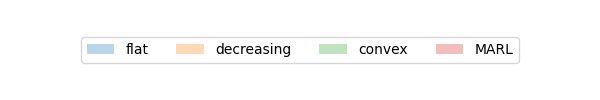}
  \includegraphics[width=0.6\linewidth]{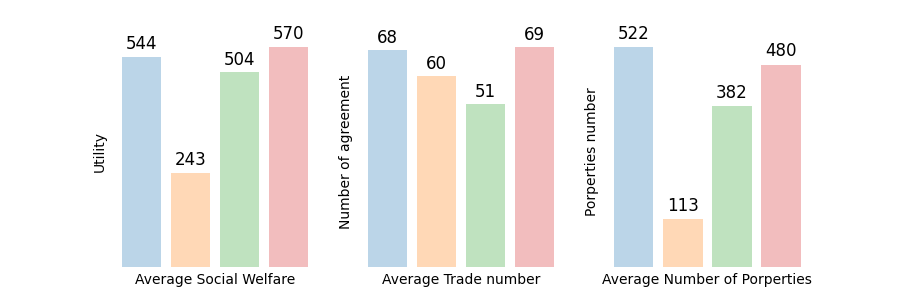}
  \includegraphics[width=0.6\linewidth]{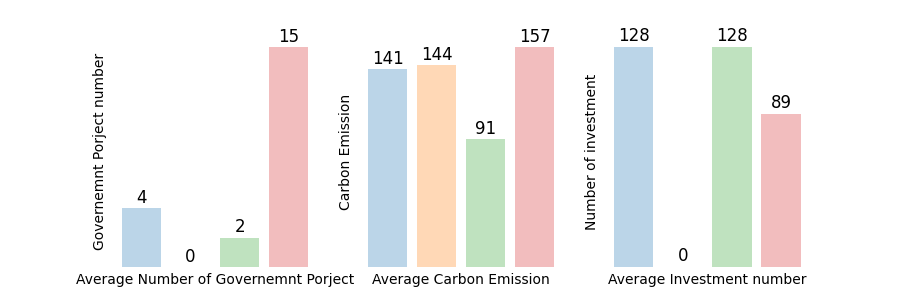}
  \caption{Simulation details for \textbf{MARL} and other approaches with \textbf{SI}, which include the social welfare, cumulative carbon emissions, the total number of trade activities, properties, government projects, and the investments.}
  \label{fig:2d}
\end{figure}

To enhance agents' performance, the observation of enterprises includes position information, which can be represented as an image. To process this position information, a convolutional neural network (CNN) is employed. Other information is added to the CNN's output, and the combined data is passed through a pipeline consisting of fully connected (FC) layers, a long short-term memory (LSTM) network, and additional FC layers. This process yields action logits. Agents' policies are obtained by applying an action mask and performing a \texttt{softmax} operation on these logits.

MARL training aims to discover an allocation policy that maximizes the Government's reward $r$, while also finding a balance between productivity, equality, and carbon emissions under the designated economy-climate coefficient $c^e$.

For the joint optimization of enterprise and government policies, we first initialize the parameters of enterprise agents to those trained under government policies based on \textbf{Flat} and Enterprise size as the indicator (\textbf{SI}) scenario. Subsequently, the parameters of government policies are randomly initialized. During training, we utilize the PPO algorithm~\cite{de2020independent} under the RLlib framework~\cite{liang2018rllib}. Additionally, we experiment with various hyperparameters for both enterprise and government agents, including learning rate and entropy regularization. Following training with $400$ million samples, we find both enterprise and government agents to converge to stable policies, which effectively balance productivity, equity, and carbon emissions. (Figure~\ref{fig:bar})

\subsection{Baseline Allocation Policies}
We utilize the indicator approach~\cite{ZHOU201647} to allocate the proportion of total carbon credits for each enterprise annually, with Emission (also called \textbf{grandfathering} or \textbf{GF}~\cite{zetterberg2012short}), Emission intensity (also called \textbf{benchmarking} or \textbf{BM}~\cite{groenenberg2002benchmark}), and Enterprise size selected as indicators (shorten as \textbf{SI}).
Due to the large temporal scale of our simulation spanning $10$ years, the government agent needs to allocate the total carbon emission credits for each year over the $10$-year period. 
We refer to the global carbon emission historical data and future forecasts provided by the IPCC~\cite{IPCC:2023} to establish emission scenarios on a large temporal scale: we call this scenario \textbf{Convex}. 
Additionally, we provide scenarios where the annual emission targets decrease gradually over time (\textbf{Decreasing}), and scenarios where the annual emission targets remain constant over time (\textbf{Flat}).    
    
\subsection{Simulation Details}
In this section, we delve into the specifics of our simulation setup and the observed behaviors of enterprise agents under various government policies. We analyze key activities such as trade, production, and the construction of government-certified projects, assessing their impact on overall productivity, equality, and carbon emissions. Through these simulations, we aim to illustrate how different allocation policies influence enterprise behavior and the resulting economic and environmental outcomes.

\paragraph{Trade Activities.}
In each \texttt{trade} activity, a carbon emission credit circulates among the enterprises, along with coins corresponding to the current price of the carbon emission credit. 
The price of emission credit fluctuates significantly with changes in enterprise carbon emission level, making it challenging to analyze. 
Therefore, we analyze enterprise trading activities by averaging trading volumes over multiple episodes. 
As shown in the graph in the middle-upper part of Figure~\ref{fig:2d}, under various government policies, the circulation of coins among enterprises due to trading aligns closely with the equality of government policies.

\paragraph{Production Activities.}
\texttt{Produce} action is the only action that generates coins, and the coins generated in a single production are solely related to the enterprise's size. 
Therefore, the total income of all enterprises in an episode is closely related to the total number of \texttt{produce} actions, which is essentially the number of properties. 

\paragraph{Government Certified project.}
Government-certified projects in our setup represent government-led public green energy initiatives, aligning with real-world scenarios.
In our simulator, these projects can store carbon credits across periods. 
When constructed, they initially reduce the carbon emissions of all enterprises (defined in Equation~\ref{invest}). 
They also pay $1$ unit of carbon emission credit to the constructor (i.e., enterprise) but charge high coins and labor costs.
As observed in the bottom-left part of Figure \ref{fig:2d}, all baseline policies exhibit lower total government-certified project construction. 
In contrast, the MARL-based allocation policy stands out with significantly higher government-certified project construction. 
This difference may stem from the MARL-based policy's more diversified allocation strategies, resulting in higher social welfare.

\paragraph{Carbon emission.}
According to Equation~(\ref{invest}), \texttt{invest} actions can reduce the carbon emissions of the investing enterprise while also having a minor impact on the carbon emissions of other enterprises. 
The overall carbon emissions are related to the overall production activities and the carbon emissions of individual enterprises.
From the lower-middle and lower-right parts of Figure~\ref{fig:2d}, in conjunction with the previous analyses of production activities and government-certified projects, we can observe different behavioral patterns associated with the $4$ different policies.
Under the \texttt{flat} policy, enterprises engage in more total production and investment activities, resulting in higher carbon emissions. 
However, due to the lack of differentiation in this policy, some enterprises exceed their allocated carbon credits significantly.
In the \texttt{decreasing} policy, economic activities of enterprises are subdued, leading to lower productivity but higher equality and fewer excess carbon emissions.
For the \texttt{convex} policy, there are fewer total production activities but more investment activities, resulting in lower and fewer excess carbon emissions.
Under the MARL-based policy, enterprises construct more government-certified projects that can enhance overall income. 
Due to differentiated allocation strategies, even with lower investment activity, there are more total production activities and higher carbon emissions but fewer excess carbon emissions.

\section{Visualization}
In this section, we present the visual tools developed to illustrate the simulation results and analyze the behaviors of the agents in our carbon market model. Visualization plays a crucial role in understanding the complex interactions and outcomes within the multi-agent reinforcement learning (MARL) environment. Our visual tools provide insights into the dynamics of enterprise and government agents, their economic activities, and the impact of different policies.

\begin{figure}[htb!]
    \centering
    \includegraphics[width=0.9\linewidth]{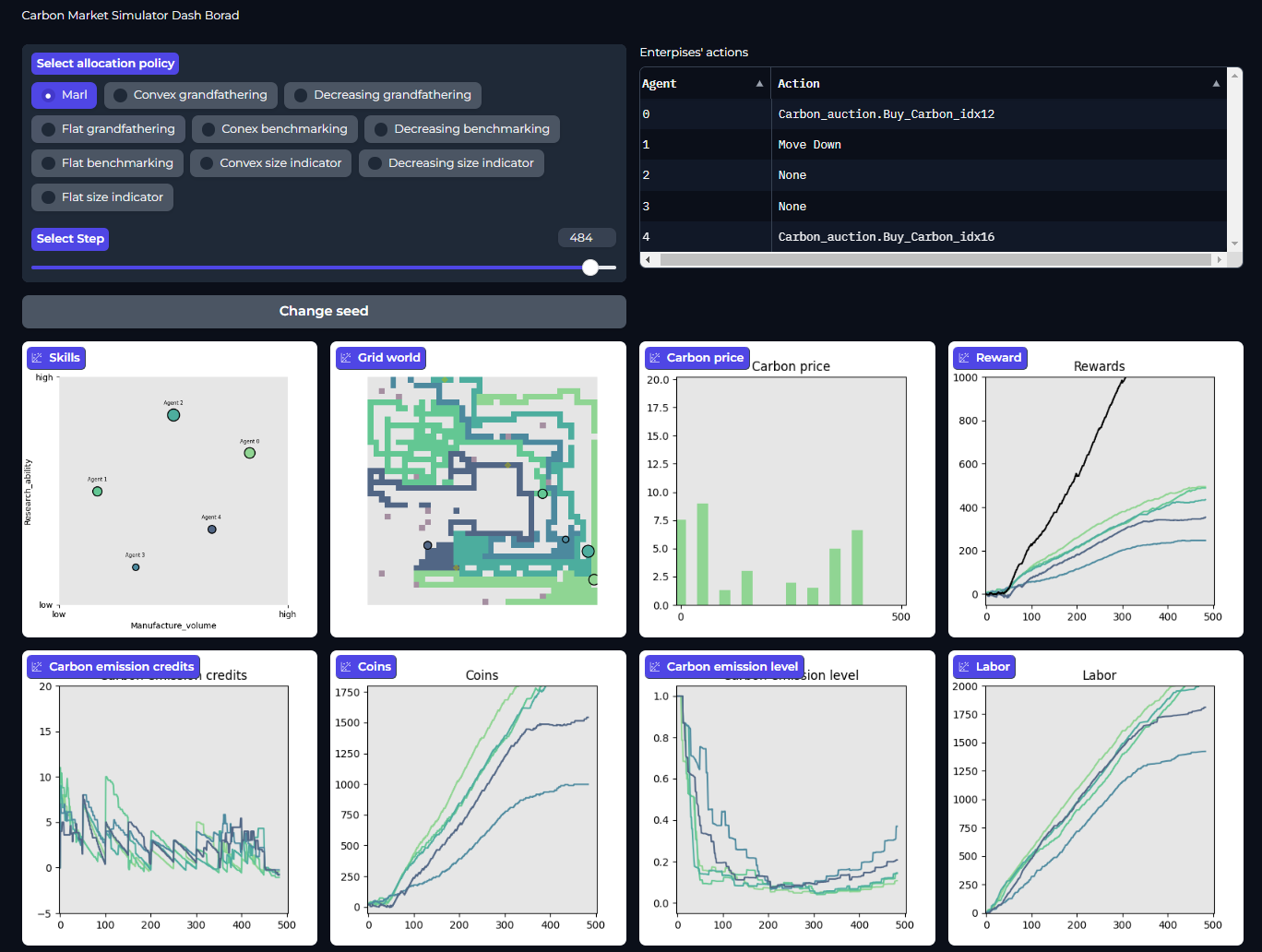}
    \caption{Simulator dashboard, it presents detailed information encompassing enterprises' attributes, assets, and actions within a single time step across various example episodes under different policies. Additionally, it provides visual representations of the average carbon prices over different periods and presents rewards for both enterprises and government.}
    \label{fig:Demo}
\end{figure} 

\subsection{Dashboard Overview}

The simulator dashboard (Figure~\ref{fig:Demo}) offers a comprehensive view of the simulation at any given time step within an episode. It includes detailed information about the attributes, assets, and actions of enterprises, as well as visual representations of the average carbon prices, trade volumes, and rewards for both enterprises and the government. This allows for an intuitive comparison between different baseline and MARL strategies.

The dashboard is divided into several key sections:
\begin{itemize}
    \item \textbf{Skills}: Displays the research ability and manufacturing volume of each enterprise.
    \item \textbf{Grid World}: Visualizes the positions and movements of enterprises within the grid world.
    \item \textbf{Carbon Price}: Tracks the fluctuations in carbon emission credit prices.
    \item \textbf{Rewards}: Shows the cumulative rewards for both enterprises and the government.
    \item \textbf{Carbon Emission Credits}: Illustrates the remaining carbon credits for each enterprise.
    \item \textbf{Coins}: Displays the accumulation of coins by each enterprise.
    \item \textbf{Carbon Emission Level}: Monitors the changes in carbon emission levels over time.
    \item \textbf{Labor}: Tracks the labor usage of each enterprise.
\end{itemize}

By leveraging these visualizations, we gain valuable insights into the trade-offs between productivity, equality, and carbon emissions, ultimately guiding the development of more effective carbon market policies.

\section{Closing Remarks}
    In this paper, enterprise and government agents participate in carbon market simulations via MARL-based adaptive mechanism design. 
    By fine-tuning the government's reward function, we can exploit the adaptability to strike a balance between various economic and climate objectives. 
    Unlike the commonly used indicator approach, MARL-based agents can incorporate more comprehensive information, enabling them to formulate more personalized and diversified allocation strategies.
    We also illustrates the practicality of employing hierarchical model-free MARL for carbon market simulation. 
    It envisions the potential of machine learning to contribute to global emission reduction endeavors. 
    However, the proposed simulator still needs to be improved, notably the absence of empirical modeling for emissions reduction investments made by enterprises. 
    Consequently, future simulations can enhance their realism by integrating more real-world data.

\section*{Ethical Statement}
    
    In our development of the carbon market simulator, we adhere to principles of transparency, integrity, and fairness, ensuring compliance with the highest ethical standards while advancing understanding in environmental economics. We prioritize privacy, equity, and social responsibility throughout our research and development process. However, it's important to acknowledge that our simulator may not encompass all aspects of the real world. As such, we do not endorse the use of learned policies for actual policy making.

\section*{Acknowledgements}

Baoxiang Wang is partially supported by National Natural Science Foundation of China (62106213, 72394361).

\newpage

\bibliographystyle{unsrt}  
\bibliography{references.bib}

\end{document}